\documentclass{article}
\usepackage{ijcai25}

\usepackage{times}
\usepackage{soul}
\usepackage{url}
\usepackage[hidelinks]{hyperref}
\usepackage[utf8]{inputenc}
\usepackage[small]{caption}
\usepackage{graphicx}
\usepackage{amsmath}
\usepackage{amsthm}
\usepackage{booktabs}
\usepackage{algorithm}
\usepackage{algorithmic}
\usepackage[switch]{lineno}
\usepackage{color}
\usepackage{subcaption}

\title{KGCompiler: Deep Learning Compilation Optimization for Knowledge Graph Complex Logical Query Answering}

\author{
    Hongyu Lin\textsuperscript{\rm 1,2}$^\ast$, Haoran Luo\textsuperscript{\rm 3}$^\ast$, Hanghang Cao\textsuperscript{\rm 1,2}\thanks{Equal contribution.}, Yang Liu\textsuperscript{\rm 1,2}, Shihao Gao\textsuperscript{\rm 1,2}, \\
    Kaichun Yao\textsuperscript{\rm 2}, Libo Zhang\textsuperscript{\rm 2}, Mingjie Xing\textsuperscript{\rm 2}\thanks{Corresponding author.}, Yanjun Wu\textsuperscript{\rm 2} \\
    \normalfont\textsuperscript{\rm 1}University of Chinese Academy of Sciences \\
    \normalfont\textsuperscript{\rm 2}Institute of Software Chinese Academy of Sciences \\
    \normalfont\textsuperscript{\rm 3}Beijing University of Posts and Telecommunications \\
    \normalfont\texttt{hongyu2021@iscas.ac.cn, luohaoran@bupt.edu.cn} \\
}

\begin{document}

\maketitle

\begin{abstract}
    Complex Logical Query Answering (CLQA) involves intricate multi-hop logical reasoning over large-scale and potentially incomplete Knowledge Graphs (KGs). Although existing CLQA algorithms achieve high accuracy in answering such queries, their reasoning time and memory usage scale significantly with the number of First-Order Logic (FOL) operators involved, creating serious challenges for practical deployment. In addition, current research primarily focuses on algorithm-level optimizations for CLQA tasks, often overlooking compiler-level optimizations, which can offer greater generality and scalability. To address these limitations, we introduce a \textbf{K}nowledge \textbf{G}raph \textbf{Compiler}, namely \textit{KGCompiler}, the first deep learning compiler specifically designed for CLQA tasks.
    By incorporating KG-specific optimizations proposed in this paper, \textit{KGCompiler} enhances the reasoning performance of CLQA algorithms without requiring additional manual modifications to their implementations. At the same time, it significantly reduces memory usage. Extensive experiments demonstrate that KGCompiler accelerates CLQA algorithms by factors ranging from 1.04$\times$ to 8.26$\times$, with an average speedup of 3.71$\times$. We also provide an interface to enable hands-on experience with \textbf{KGCompiler}~\footnote{Github Code: \url{https://github.com/LHY-24/KG-Compilation}}.
\end{abstract}

\section{Introduction}
Knowledge Graphs (KGs) represent real-world knowledge in the form of head-relation-tail triples, which have been applied in various fields, such as biomedical knowledge system \cite{himmelstein2017systematic}, recommender systems \cite{wang2018dkn} and common sense reasoning \cite{lv2020graph}. \par

\begin{figure}[t]
\centering
\includegraphics[width=80mm]{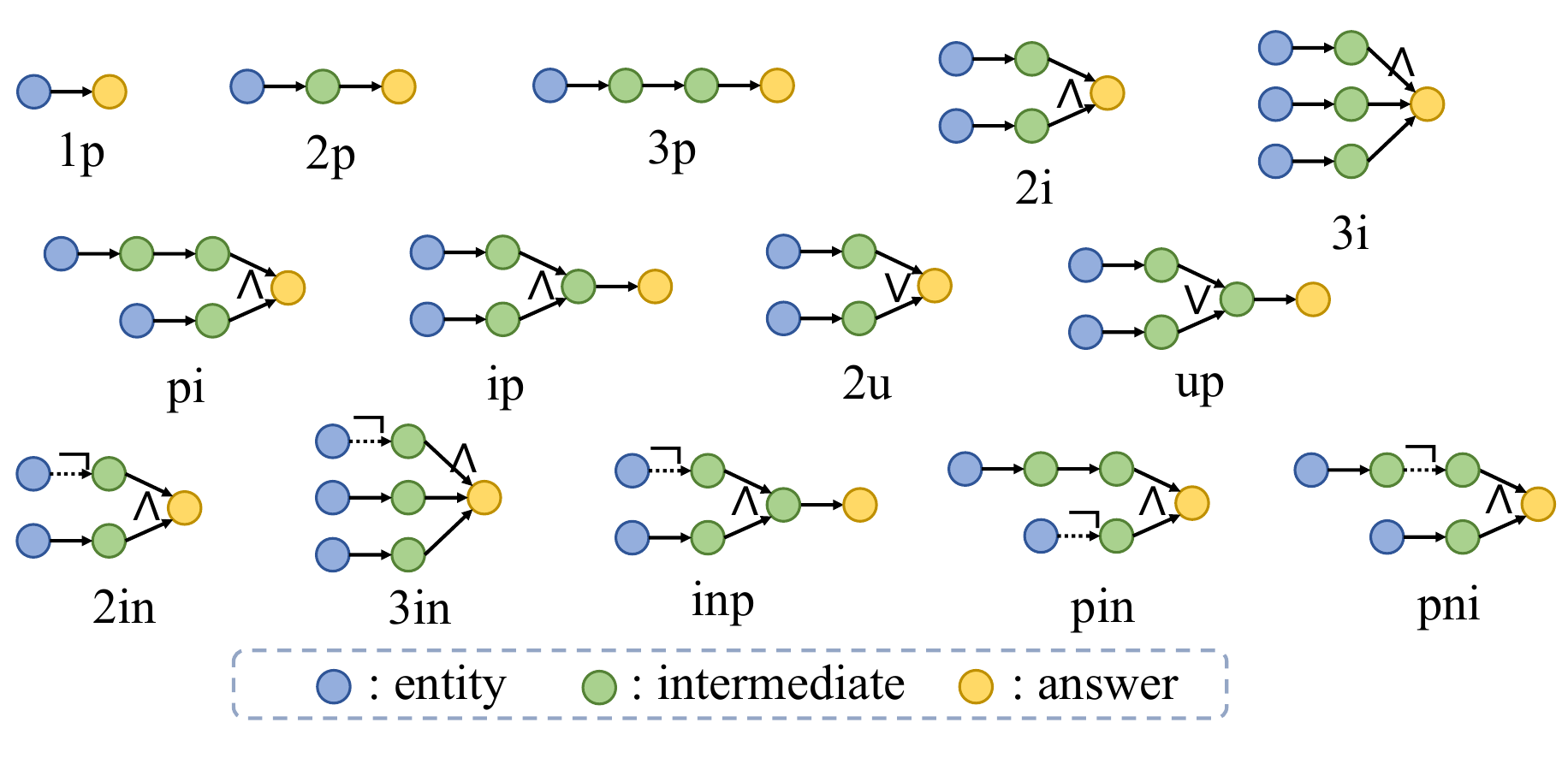}
\caption{An illustration of the different types of CLQA tasks. Symbol explanation: $\land$ represents intersection, $\lor$ represents union, $\neg$ represents negation, and $\rightarrow$ represents projection. FOL operators naming convention: $p$ - projection, $i$ - intersection, $n$ - negation, $u$ - union.}
\label{queries}
\end{figure}

\begin{figure}[t]
\centering
\includegraphics[width=0.45\textwidth]{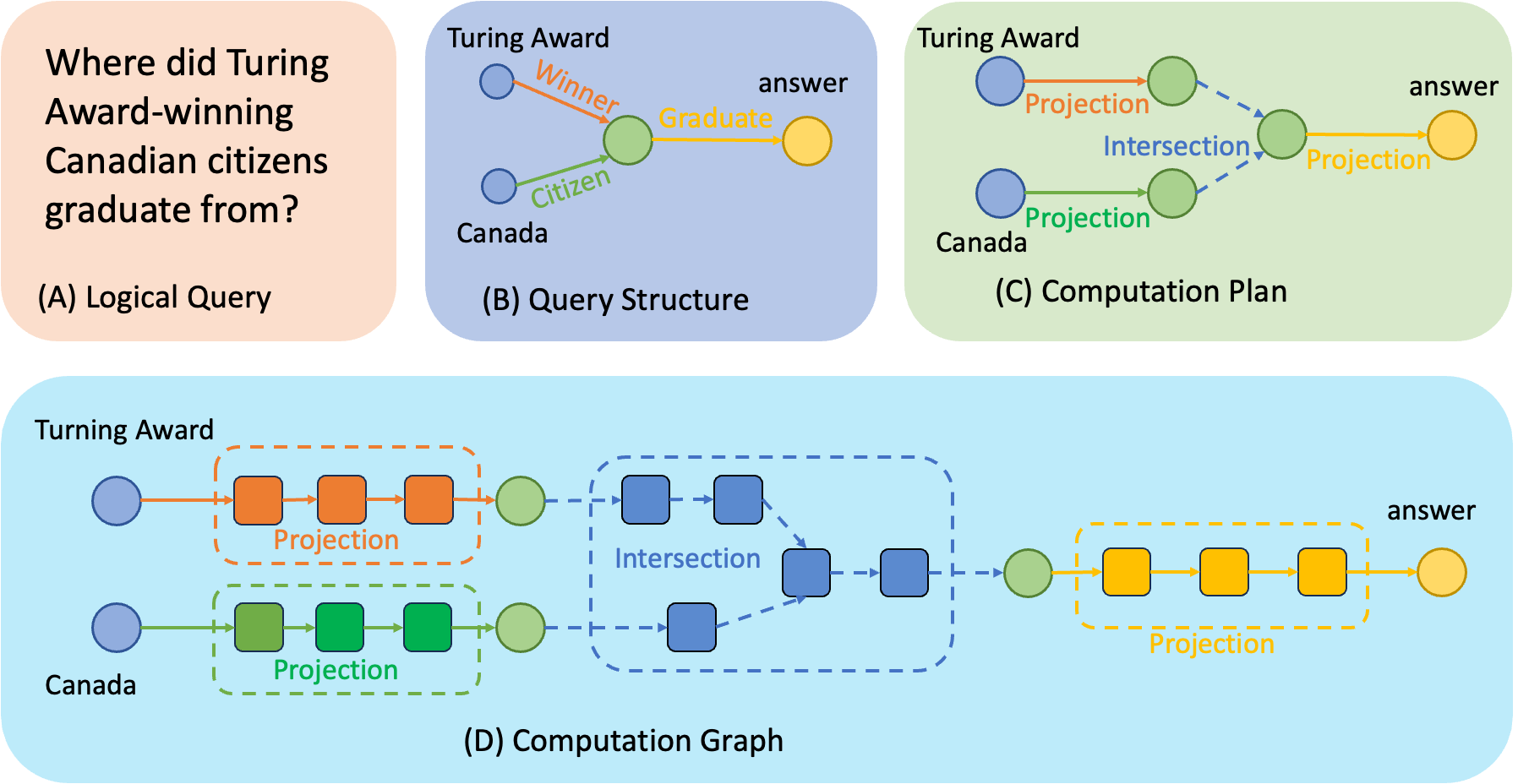}
\caption{Overview of the FOL query process mapped to the Computation Graph: The complex Logical Query (A) is first parsed into a Query Structure (B). Subsequently, the query Computation Plan (C) is followed to execute the query directly within the Computation Graph (D).}
\label{introduction}
\end{figure}

Complex Logical Query Answering (CLQA) is one of the most representative types of reasoning tasks, which is concerned with answering complicated queries over KGs. As illustrated in Figure \ref{queries}, CLQA tasks involve composite structures connecting multiple nodes, utilizing four key First-Order Logic (FOL) operators: conjunction ($\land$), disjunction ($\lor$), negation ($\neg$), and existential quantifier ($\exists$). These operations enable the solution of multi-hop logical reasoning tasks over large-scale, even potentially incomplete KGs. For example, a complex logical query such as “Where did Canadian citizens with Turing Award graduate?” can be formulated as the corresponding FOL query shown in Figure \ref{introduction}. \par

However, the current CLQA domain faces three significant challenges in practical applications. \textbf{Challenge 1: Multi-hop tasks lead to longer reasoning time.} The reasoning time for CLQA tasks is proportional to the number of FOL operators involved. For instance, the $3in$ task requires substantially more reasoning time than the $1p$ task. As the complexity of the task increases, the required reasoning time also grows, negatively impacting the efficiency of KG models. This challenge is prevalent across almost all CLQA algorithms. \textbf{Challenge 2: Complex tasks demand larger memory footprint.} More complex CLQA tasks typically involve deeper inference paths and more query branches, resulting in larger memory footprints. However, the limited memory capacity of computing devices restricts the batch size during each iteration, hindering the use of large-scale parallelism to accelerate reasoning in KG models. \textbf{Challenge 3: Lack of general compilation optimizations for the CLQA tasks.} Existing research mainly focuses on algorithm-level innovations to enhance the accuracy of CLQA tasks and pays less attention to compiler-level optimizations. Compilation optimizations, however, offer more general and scalable solutions for improving the overall performance of KG models. \par

To address these challenges, we propose KG-oriented compiler-level optimizations and integrate them into a \textbf{K}nowledge \textbf{G}raph \textbf{Compiler} (\textit{KGCompiler}), the first KG-oriented deep learning compiler specifically designed for CLQA tasks. \textit{KGCompiler} consists of three core components: (1) Graph Capturer: Converts KG models into their corresponding computation graph representations. (2) Pattern Recognizer: Detects specific combinations of FOL operators within the computation graph. (3) Operator Fuser: Applies KG-oriented optimizations to the computation graph based on the identified patterns. By leveraging these components, KGCompiler effectively addresses the challenges outlined earlier: (1)\textbf{Long Reasoning Time}: \textit{KGCompiler} significantly enhances the reasoning performance of leading KG models across 14 classic CLQA tasks. (2) \textbf{Large Memory Footprint}: It reduces memory usage by increasing data reuse and minimizing the number of kernels required for computation. (3) \textbf{General Compilation Optimizations}: \textit{KGCompiler} automatically applies tailored optimizations specific to each KG model's unique characteristics, eliminating the need for manual intervention.\par

To evaluate the performance of \textit{KGCompiler}, we compare it against the prior CLQA-oriented framework KGReasoning \cite{kgreasoning}, using six representative CLQA algorithms and three standard benchmark datasets. Specifically, we assess performance on 14 classic CLQA tasks proposed by \cite{NEURIPS2020_e43739bb}, including 9 Existential Positive First-Order (EPFO) queries (containing only $\exists$, $\land$, and $\lor$) ($1p$, $2p$, $3p$, $2i$, $3i$, $pi$, $ip$, $2u$, $up$) and 5 types of queries with negation ($2in$, $3in$, $inp$, $pin$, $pni$). Extensive experimental results demonstrate that \textit{KGCompiler} consistently improves the reasoning performance of all evaluated CLQA algorithms on these tasks compared to KGReasoning, achieving speedups ranging from 1.04$\times$ to 8.26$\times$, with an average improvement of 3.71$\times$. Furthermore, \textit{KGCompiler} significantly reduces the memory usage of KG models, enabling the use of larger batch sizes to accelerate reasoning. Finally, we analyze the underlying factors driving these improvements, highlighting the universality and versatility of \textit{KGCompiler}. \par

\section{Related Work}
\paragraph{Complex Logical Query Answering.} Numerous CLQA-related algorithms, such as GQE \cite{hamilton2018embedding}, Q2B \cite{ren2020query2box}, BetaE \cite{NEURIPS2020_e43739bb}, CQD \cite{arakelyan2020complex}, LogicE \cite{LogicE}, ConE \cite{ConE}, GNN-QE \cite{zhu2022neural}, QTO \cite{bai2023answering}, Query2Triple \cite{Q2T}, and FIT \cite{yin2024rethinking}, have been developed to achieve more accurate query answers by leveraging various entity representations and reasoning strategies. Frameworks such as KGReasoning \cite{kgreasoning} and SMORE \cite{ren2021smore} provide essential infrastructures for training and evaluating KG models, thereby facilitating the integration and application of various CLQA algorithms. For the first time, we propose CLQA-oriented compilation optimizations in this paper, which introduce a novel perspective for optimizing KG models. Unlike previous works that primarily focus on improving the accuracy of query answers, our optimizations aim to enhance reasoning performance and memory footprint of KG models. This distinction highlights the unique contributions of our approach to the field. \par

\paragraph{Deep Learning Compilation.}
Deep Learning Compilation transforms the deep learning models from various deep learning frameworks, such as PyTorch \cite{paszke2019pytorch} and Tensorflow \cite{abadi2015tensorflow}, into optimized executable kernels for diverse deep learning hardware backends, including CPUs, GPUs, TPUs, Wafer-scale chip \cite{wafer}, etc. Several deep learning compilers, such as TVM \cite{chen2018tvm}, Hidet \cite{ding2023hidet}, and BladeDISC \cite{10.1145/3617327}, have been proposed by both industry and academia, significantly enhancing the performance of deep learning models through multi-layer intermediate representations and corresponding optimizations. While most existing deep learning compilers focus on optimizing deep neural network models, we extend their functionality by incorporating the unique characteristics of KG models. This extension achieves notable improvements in non-neural network applications, such as CLQA tasks. \par

\section{Preliminaries}
\paragraph{Knowledge Graph.} A knowledge graph (KG) can be formally represented as $\mathcal{G = (\xi,R,T)}$. Here, $\xi = \{e_1,e_2,...,e_n\}$ denotes the entity set, where each $e_i \in \mathcal{\xi}$ represents a constant or variable entity in the real world. $\mathcal{R} = \{r_1,r_2,...,r_m\}$ denotes the relation set, where each relation $r_i \in \mathcal{R}$ captures the semantic relationship between a pair of entities. Finally, $\mathcal{T} = \{(h,r,t) | h,t \in \xi, r \in \mathcal{R}\}$ represents the set of triples, with each triple $(h,r,t)$ encoding a fact from head entity $h$ to tail entity $t$ with relation $r$. \par

\paragraph{Complex Logical Query Answering.} CLQA focuses on answering first-order logic (FOL) queries over KGs, which involve multi-hop reasoning containing four logical operators: conjunction ($\land$), disjunction ($\lor$), negation ($\neg$), and existential quantifier ($\exists$). Specifically, a FOL query $Q$ consists of a set of non-variable anchor entities $\xi_Q \subseteq \xi$, a set of existentially quantified bound variables $V_Q = \{v_1,...,v_n\} \subset \mathcal{\xi}$, and the answer to $Q$ is the set of viable assignments to the target entity variable $v_? \in \{v_1,...,v_n\}$ \cite{NEURIPS2020_e43739bb}. The disjunctive normal form (DNF) of $Q$ can be expressed as Equation \ref{eq:FOL}, where each clause $c_i$ denotes a conjunction of a series of literals $l_{ij}$. Each literal is either an atomic formula or its negation, composed of an entity $e \in \xi_Q$, a variable $v \in \{v_?, v_1,...,v_n\}$, and a variable $v' \in \{v_1,...,v_n\}$. \par

\begin{equation}
\label{eq:FOL}
  \begin{split}
  Q[v_?] &= v_? . \exists v_1,...,v_k : c_1 \vee c_2 \vee ... \vee c_n, \\
  c_i &= l_{i1} \wedge l_{i2} \wedge ... \wedge l_{in}, \\
  l_{ij} &= r(e,v) \text{ or } \neg r(e,v) \text{ or } r(v',v) \text{ or } \neg r(v',v).
  \end{split}
\end{equation}

For example, a FOL query "Where did Turing Award-winning Canadian citizens graduate from?" can be represented as: $Q[v_?] = v_? . \exists v : Winner(Turing Award, v) \wedge Citizen(Canada,v) \wedge Gradudate(v,v_?)$. 

\paragraph{Query Computation Graph.} A query computation graph in KGCompiler is represented as a bipartite directed acyclic graph $G = (V,E,F)$. The node set is partitioned into two disjoint sets: $V = V_{var} \cup V_{op}$, where $V_{var} \cap V_{op} = \emptyset$. Specifically, $V_{var} = \{var_1,var_2,...,var_n\}$ is the set of value nodes, which denotes the inputs, outputs, and intermediate results of the computation process. On the other hand, $V_{op} = \{op_1, op_2,...,op_n\}$ is the set of operator nodes, representing the primitive deep learning operations involved in the computation. The edge set $E = \{(v_i,v_j) | v_i,v_j \in V\}$ consists of directed edges that capture data dependencies between the nodes, where the output of node $v_i$ serves as the input to node $v_j$. The set $F$ contains functions, where each function $f_i \in F$ defines a specific relationship within the computation graph. \par

\section{Methodology}

\begin{figure*}[htbp]
\centering
\includegraphics[width=0.85\textwidth]{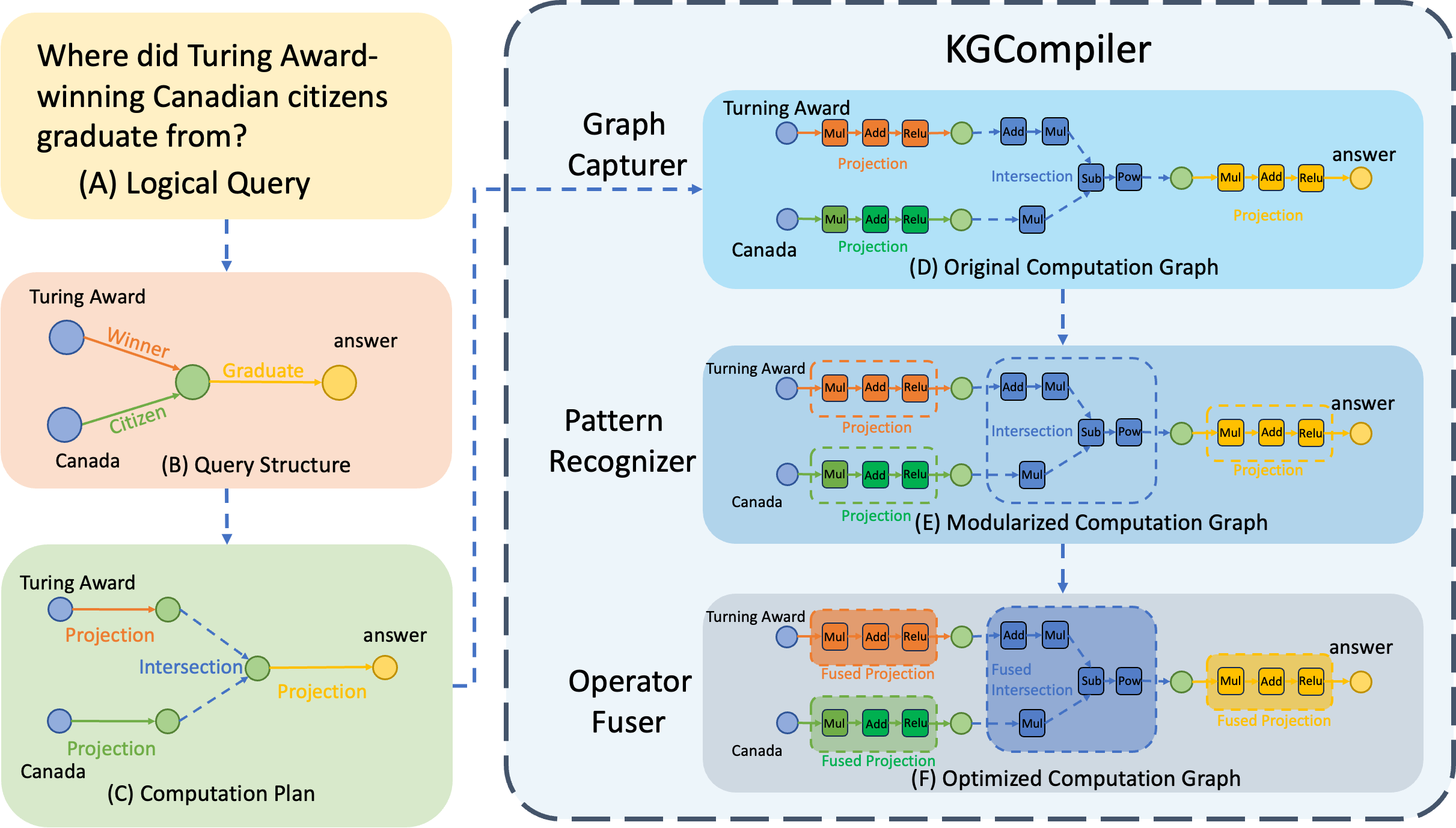}
\caption{Architecture of \textit{KGCompiler}. To begin with, the Logical Query (A) is parsed into a Query Structure (B), which is then mapped to a Computation Plan (C). Next, Graph Capturer processes this Computation Plan to construct the Original Computation Graph (D). Following this, Pattern Recognizer extracts the combination patterns in the graph, resulting in a Modularized Computation Graph (E). Finally, by applying the fusion optimizations based on the recognized patterns, Operator Fuser generates the Optimized Computation Graph (F).}
\label{architecture}
\end{figure*}

\subsection{Overview of KGCompiler}

In this section, we present \textit{KGCompiler}, a deep learning compiler designed specifically for CLQA tasks on KGs. \textit{KGCompiler} consists of three key components: the Graph Capturer, the Pattern Recognizer, and the Operator Fuser. The architecture of KGCompiler is shown in Figure \ref{architecture}.

While not all CLQA tasks rely on deep neural networks, it is worth noting that the data flow in most CLQA-related KG models can still be represented by computation graphs. In this paper, we are the first to propose leveraging deep learning compilation optimizations within the domain of KG. By converting KG models into their corresponding computation graphs and applying a series of KG-specific, graph-level deep learning compilation optimizations, we demonstrate significant improvements in the performance of KG models. Unlike prior KG-oriented optimizations, the compilation optimizations introduced in this paper are directly applicable to a wide range of KG models without requiring manual modifications to their implementations, reflecting the generality and usability of our approach. Furthermore, these compilation optimizations are orthogonal to existing techniques, allowing them to be seamlessly integrated with other methods to provide synergistic enhancements for KG models. 


\subsection{Graph Capturer}
As we mentioned above, the computation graph serves as the foundation for optimizations within the KGCompiler. To facilitate this, we introduce Graph Capturer $M$, which converts a FOL query $Q$ into its corresponding computation graph $G$. Specifically, each entity $e_i \in \xi_Q$ and variable $v_j \in \{v_1,v_2,...,v_?\}$ in $Q$ are first mapped to value nodes $var_i,var_j \in V_{var}$, separately. Each relation $r_k(e_i,v_j) \in \mathcal{R}$ is then converted into a series of consecutive edges $\{(var_i,op_k), (op_k,var_j)\} \subset E$, where $op_k \in V_{op}$ represents the operation node associated with the relation. Additionally, the four FOL operators ($\land$, $\lor$, $\neg$, $\exists$) between subqueries $Q_1,Q_2$ are mapped to the function set $F = \{f_{and},f_{or},f_{not},f_{project}\}$. Consequently, the entire data flow of $Q$ on the KG is represented by $G$, forming the foundation for the subsequent pattern recognition module. The whole converting process can be formulated as below:

\begin{equation}
\label{eq:mapping}
G = M[Q] = \left \{
\begin{aligned}
    &e_i \rightarrow var_i, \\
    &v_j \rightarrow var_j, \\
    &r_k(e_i, v_j) \rightarrow \{(var_i,op_k), (op_k,var_j)\}, \\
    &\land(Q_1,Q_2) \rightarrow f_{and}(M[Q_1],M[Q_2]), \\
    &\lor(Q_1,Q_2) \rightarrow f_{or}(M[Q_1],M[Q_2]), \\
    &\neg(Q) \rightarrow f_{not}(M[Q]), \\
    &\exists x.Q \rightarrow f_{project}(M[Q],x).
\end{aligned}
\right.
\end{equation}

\subsection{Pattern Recognizer}
Due to the diverse approaches employed by various algorithms to address the CLQA tasks, the embeddings of entities and variables in corresponding KG models are different. Adding to the difficulty, the implementation of FOL operators (projection $(p)$, intersection $(i)$, union $(u)$, negation $(n)$) also varies across these models. Therefore, we propose Pattern Recognizer $P$ to automatically identify the combination patterns of these operators within $G$, enabling the application of model-specific optimizations tailored to their unique characteristics. \par

For instance, BetaE \cite{NEURIPS2020_e43739bb} learns a Beta embedding $Beta(x;\alpha_i,\beta_i)$ for each entity or variable $e_i \in \mathcal{\xi}$, where $x \in [0,1]$ represents the domain of the Beta function, and $\alpha_i,\beta_i$ are the shape parameters. As a result, a given query $Q$ in BetaE will be represented as a collection of  independent Beta distributions $S = [(\alpha_1,\beta_1),...,(\alpha_n,\beta_n)]$. Formally,\par

\begin{equation}
\label{eq:beta}
\begin{aligned}
\text{Beta}(x; \alpha, \beta) = \frac{x^{\alpha-1} (1-x)^{\beta-1}}{B(\alpha, \beta)}  \\
B(\alpha, \beta) = \int_0^1 t^{\alpha-1} (1-t)^{\beta-1} \, dt.
\end{aligned}
\end{equation}

When it comes to the FOL operators, for example, BetaE define a probabilistic projection operator $p$ that maps one Beta embedding $S$ to another Beta embedding $S'$ given the relation type $r$, which can be formulated as a multi-layer perceptron (MLP), where $S' = \text{MLP}_r(S)$. The Pattern Recognizer $P$ automatically extracts the combination patterns of $\text{MLP}_r$ within $G$, which can be decomposed into a sequence of primitive deep learning operators and variables, as shown in Equation \ref{eq:MLP}. While $\{op_{matmul}, op_{add}, op_{relu}, op_{softmax}\} \subset V_{op}$ is the set of primitive operators involved, and $\{var_S,var_{S'},var_{w1},...,var_{t1},...,var_{b1},...\} \subset V_{var}$ is the set of values corresponds to the embeddings, weights, temporary results, and bias in the KG model. \par

\begin{equation}
\label{eq:MLP}
P[\text{MLP}_r(S)] \equiv \left\{
\begin{aligned}
    & \quad var_{t1} = op_{matmul}(var_S, var_{w1}), \\
    & \quad var_{t2} = op_{add}(var_{t1}, var_{b1}), \\
    & \quad var_{t3} = op_{relu}(var_{t2}), \\
    & \quad var_{t4} = op_{matmul}(var_{t3},var_{w2}), \\
    & \quad var_{t5} = op_{add}(var_{t4}, var_{b2}) \\
    & \quad var_{t6} = op_{relu}(var_{t5}), \\
    & \quad var_{t7} = op_{matmul}(var_{t6}, var_{w3}), \\
    & \quad var_{t8} = op_{add}(var_{t7}, var_{b3}), \\
    & \quad var_{S'} = op_{softmax}(var_{t8}). \\
\end{aligned}
\right.
\end{equation}

In Figure \ref{architecture} (E), a FOL operator is expressed by a rectangle, which corresponds to a specific combination pattern of primitive deep learning operators. Based on these patterns, \textit{KGCompiler} can automatically determine the fusible operators within $G$, thereby guiding further fusion optimization. \par

\subsection{Operator Fuser}
Operator fusion is an crucial technique for optimizing computation graphs. It involves merging multiple operators into a single, more efficient hardware kernel without changing the overall output. A well-defined fusion strategy can significantly improve computational efficiency, reduce memory usage, and minimize kernel launch overhead. Based on the combination patterns of FOL operators in $G$, we propose KG-oriented operator fusion strategies, which can cover the major combination patterns of FOL operators. \par

\paragraph{Horizontal Fusion Strategy.} Merge operators within the same query path as much as possible. The FOL operators $p$ and $n$ typically involve sequential operations on a single query path (Figure \ref{queries}). To optimize this, we propose fusing multiple consecutive $p$ operators into a single kernel, thereby eliminate redundant intermediate steps or temporary variables. This optimization is particularly effective for tasks such as $2p$ and $3p$. \par

\paragraph{Vertical Fusion Strategy.} Merge operators across different query paths. FOL operators $i$ and $u$ often span multiple query paths (Figure \ref{queries}). To optimize this, we propose fusing the operators that span different paths into a single kernel, where intermediate results from one path are integrated with operations from another path, thereby reducing redundant intermediate outputs and computational steps across different paths. This optimization works well for tasks such as $2i$, $3i$, $2u$, etc. \par

\paragraph{Hybrid Fusion Strategy.} A combination of both horizontal and vertical fusions. Specifically, for cases involving the integration of these two types of operators, we suggest fusing operators that occur both within the same path and across multiple paths in a more efficient and flexible manner. This strategy aims to strike a balance between local and global optimal, which proves effective for tasks such as $pi$, $ip$, $up$, $2in$, $3in$, $inp$, $pin$, $pni$, etc. \par

Figure \ref{architecture} illustrates how fusion strategies perform in Operator Fuser, which can be mainly divided into the three steps, as outlined in Algorithm \ref{alg:algorithm}. \par
\paragraph{Step1: Determine each operators' fusion strategy.} Based on the characteristic of each FOL operator in $G$, we can determine its corresponding fusion strategy. Operators with the same strategy are collected into a group, which means they will be fused in the following steps. \par
\paragraph{Step2: Apply fusion strategies.} By merging the functions and relations of each operator within a fusion group, we can ensure that the fused operators perform the same functions as all the operators in the group while maintaining the outer dependencies of them. \par
\paragraph{Step3: Update the computation graph.} Once all the fused operators have been generated, the original operators in $G$ are replaced with their corresponding fused operators, resulting in the optimized computation graph $G_{fused}$. \par 

\begin{algorithm}[htbp]
    \caption{Operator Fuser Algorithm}
    \label{alg:algorithm}
    \textbf{Input}: Modularized Computation Graph $G(V,E,F)$, Operator Fusion Strategy List $F_S = [HorizontalFusion, VerticalFusion, HybridFusion]$. \\
    \textbf{Output}: Optimized Computation Graph $G_{fused}$. \\
    \begin{algorithmic}[1] 
        \STATE 
        \text{// Step 1: Determine the fusion strategy for each operator.}
        \\
        \FOR {each operator $op \in V_{op}$}
            \STATE $op.features$ = ExtractFeatures($op$, $G$)
            \STATE $op.strategy$ = DetermineStrategy($F_S$, $op.features$)
            \STATE $fusion\_groups$ = CollectGroups($op.strategy$, $G$)
        \ENDFOR
        \STATE
        \STATE
        \text{// Step 2: Apply fusion based on the determined strategies.} 
        \\
        \STATE $new\_ops$, $new\_edges$, $new\_functions$ = [], [], []
        \FOR {each group $g$ in $fusion\_groups$:}
            \STATE $fused\_op.func$ = MergeFunctions([$op.func$ for $op$ in $g$])
            \STATE $new\_functions$.add($fused\_op.func$)
            \STATE $fused\_op.rel$ = MergeRelations([$op.rel$ for $op$ in $g$])
            \STATE $new\_edges$.add($fused\_op.rel$)
            \STATE $fused\_op$ = CreateFusedOperator($g$)
            \STATE $new\_ops$.add($fused\_op$)
        \ENDFOR
        \STATE
        \STATE
        \text{// Step 3: Update computation graph with fused operators.}
        \\
        \STATE $G_{fused}$ = UpdateGraph($G$, $new\_ops$, $new\_edges$, $new\_functions$)
        \STATE \textbf{return} $G_{fused}$
    \end{algorithmic}
\end{algorithm}
\vspace{-5mm}

\subsection{Implementation}
Currently, we implement \textit{KGCompiler} by integrating various KG-oriented compilation optimizations on the basis of the open-source deep learning compiler TorchInductor \cite{ansel2024pytorch}. Specifically, we modify its operator resolve module and graph rewrite module by registering new FOL operators and corresponding fusion strategies, thus significantly improve its performance on CLQA tasks. \par

\begin{table*}[!ht]
    \centering
    \resizebox{1\textwidth}{!}{
    \begin{tabular}{l|c|c|ccccccccc|ccccc}
        \toprule[1pt]
        \textbf{Model} & \textbf{Framework} & \textbf{AVG$_ {speedup}$} & \textbf{1p} & \textbf{2p} & \textbf{3p} & \textbf{2i} & \textbf{3i} & \textbf{pi} & \textbf{ip} & \textbf{2u} & \textbf{up}  & \textbf{2in} & \textbf{3in} & \textbf{inp} & \textbf{pin} & \textbf{pni} \\
        \toprule[1pt]
        \multicolumn{16}{c}{\textbf{FB15K-237}} \\
        \midrule
            GQE
            & KGReasoning & \textbf{3.30$\times$} & 1.31 & 0.60 & 0.54 & 1.10 & 1.28 & 1.05 & 0.96 & 0.57 & 0.79 &-&-&-&-&-
            \\
            & KGCompiler &  & \textbf{0.17}& \textbf{0.18}& \textbf{0.20}& \textbf{0.41}& \textbf{0.47}& \textbf{0.46}& \textbf{0.43}& \textbf{0.16}& \textbf{0.28} &-&-&-&-&-
            \\
        \midrule
            Q2B
            & KGReasoning & \textbf{2.18$\times$} &  4.62 & 4.72 & 4.73 & 4.79 & 4.77 & 4.79 & 4.76 & 5.07 & 5.09  &-&-&-&-&-
            \\
            & KGCompiler & & \textbf{2.18} & \textbf{2.09} & \textbf{2.23} & \textbf{2.25} & \textbf{2.13} & \textbf{2.25} & \textbf{2.24} & \textbf{2.25} & \textbf{2.24} &-&-&-&-&-
            \\
        \midrule
            BetaE
            & KGReasoning & \textbf{7.40$\times$} & 19.46 & 18.70 & 20.24 & 20.22 & 20.29 & 18.82 & 19.31 & 19.31 & 19.82 & 22.68 & 22.43 & 22.48 & 20.94 & 20.83 \textbf{}
            \\
            & KGCompiler &  & \textbf{2.64} & \textbf{2.69} & \textbf{2.74} & \textbf{2.71} & \textbf{2.73} & \textbf{2.68} & \textbf{2.63} & \textbf{2.84} & \textbf{3.02} & \textbf{2.83} & \textbf{2.87} & \textbf{2.75} & \textbf{2.75} & \textbf{2.73}
            \\
        \midrule
            LogicE
            & KGReasoning & \textbf{1.35$\times$} & 1.33 & 1.71 & 2.03 & 2.04 & 2.31 & 2.18 & 2.36 & 1.16 & 1.6 & 2.01 & 2.25 & 2.22 & 2.25 & 2.29
            \\
            & KGCompiler &  & \textbf{1.21} & \textbf{1.41} & \textbf{1.51} & \textbf{1.56} & \textbf{1.57} & \textbf{1.47} & \textbf{1.67} & \textbf{1.18} & \textbf{1.24} & \textbf{1.4} & \textbf{1.61} & \textbf{1.52} & \textbf{1.5} & \textbf{1.57}
            \\
        \midrule
            ConE
            & KGReasoning & \textbf{6.19$\times$} & 9.79 & 10.29 & 10.75 & 11.36 & 11.98 & 11.69 & 12.01 & 16.73 & 17.25 & 11.62 & 12.15 & 12.31 & 12.49 & 12.22
            \\
            & KGCompiler &  & \textbf{0.77} & \textbf{0.9} & \textbf{0.99} & \textbf{3.14} & \textbf{3.29} & \textbf{3.29} & \textbf{3.49} & \textbf{1.15} & \textbf{1.38} & \textbf{5.11} & \textbf{6.65} & \textbf{5.82} & \textbf{6.08} & \textbf{5.96}
            \\
        \midrule
            Query2Triple
            & KGReasoning & \textbf{1.05$\times$} & 2.22 & 9.74 & 9.90 & 9.70 & 10.34 & 9.75 & 10.20 & 3.57 & 19.58 & 9.88 & 10.08 & 9.86 & 9.73 & 9.70
            \\
            & KGCompiler &  & \textbf{2.19}& \textbf{9.28}& \textbf{9.31}& \textbf{9.24}& \textbf{9.24}& \textbf{9.37}& \textbf{9.40}& \textbf{3.79}& \textbf{17.95}& \textbf{9.60}& \textbf{9.70}& \textbf{9.40}& \textbf{9.45}& \textbf{9.34}
            \\
        
        \toprule[1pt]
        \multicolumn{16}{c}{\textbf{FB15K}} \\
        \toprule
            GQE
            & KGReasoning & \textbf{3.51$\times$} & 1.88 & 0.39 & 0.54 & 0.92 & 1.04 & 1.03 & 1.18 & 0.37 & 0.71 &-&-&-&-&-
            \\
            & KGCompiler &  & \textbf{0.15}& \textbf{0.17}& \textbf{0.21}& \textbf{0.39}& \textbf{0.45}& \textbf{0.44}& \textbf{0.48}& \textbf{0.15}& \textbf{0.27} &-&-&-&-&-
            \\
         \midrule
            Q2B
            & KGReasoning & \textbf{2.14$\times$} & 4.60 & 4.73 & 4.73 & 4.79 & 4.78 & 4.54 & 4.34 & 4.92 & 5.11  &-&-&-&-&-
            \\
            & KGCompiler & & \textbf{2.19} & \textbf{2.10} & \textbf{2.25} & \textbf{2.22} & \textbf{2.26} & \textbf{2.26} & \textbf{2.27} & \textbf{2.27} & \textbf{2.14} &-&-&-&-&-
            \\
        \midrule
            BetaE
            & KGReasoning & \textbf{7.15$\times$} & 18.82 & 17.59 & 17.66 & 17.64 & 19.12 & 19.15 & 17.86 & 20.32 & 18.81 & 19.68 & 22.56 & 21.32 & 21.29 & 21.30 
            \\
            & KGCompiler &  & \textbf{2.65} & \textbf{2.62} & \textbf{2.77} & \textbf{2.62} & \textbf{2.65} & \textbf{2.78 }&\textbf{ 2.77} &\textbf{ 2.73} & \textbf{2.78} & \textbf{2.73} & \textbf{2.75} &\textbf{ 2.78} & \textbf{2.77} & \textbf{2.78}
            \\
        \midrule
            LogicE
            & KGReasoning & \textbf{1.38$\times$} & 1.12 & 1.2 & 1.7 & 1.75 & 2.11 & 2.24 & 2.16 & 1.04 & 1.39 & 1.67 & 2.23 & 1.88 & 2.2 & 2.21
            \\
            & KGCompiler &  & \textbf{0.92} & \textbf{0.97} & \textbf{1.1} & \textbf{1.23} & \textbf{1.43} & \textbf{1.49} & \textbf{1.45} & \textbf{1.09} & \textbf{1.09} & \textbf{1.32} & \textbf{1.42} & \textbf{1.44} & \textbf{1.46} & \textbf{1.44}
            \\
        \midrule
            ConE
            & KGReasoning & \textbf{6.12$\times$} & 10.03 & 10.51 & 10.94 & 11.51 & 12.18 & 12.23 & 12.02 & 17.23 & 17.72 & 12.02 & 12.79 & 12.59 & 12.41 & 12.39
            \\
            & KGCompiler &  & \textbf{1.13} & \textbf{0.94} & \textbf{1.04} & \textbf{3.16} & \textbf{3.13} & \textbf{3.16} & \textbf{3.29} & \textbf{1.13} & \textbf{1.31} & \textbf{5.03} & \textbf{5.84} & \textbf{5.62} & \textbf{5.61} & \textbf{5.64}
            \\
        \midrule
            Query2Triple
            & KGReasoning & \textbf{1.08$\times$} & 2.27 & 10.14 & 10.46 & 10.38 & 10.11 & 10.54 & 10.34 & 3.91 & 19.83 & 9.92 & 10.17 & 10.20 & 10.72 & 10.05
            \\
            & KGCompiler &  & \textbf{2.06}& \textbf{9.29}& \textbf{9.32}& \textbf{9.31}& \textbf{9.37}& \textbf{9.38}& \textbf{9.38}& \textbf{3.80}& \textbf{18.48}& \textbf{9.61}& \textbf{9.76}& \textbf{9.61}& \textbf{9.54}& \textbf{9.64}
            \\
            
        \toprule[1pt]
        \multicolumn{16}{c}{\textbf{NELL}} \\
        \midrule
            GQE
            & KGReasoning & \textbf{4.59$\times$} & 1.25 & 0.80 & 1.07 & 1.56 & 1.36 & 1.69 & 3.32 & 0.80 & 2.38 &-&-&-&-&-
            \\
            & KGCompiler &  & \textbf{0.17}& \textbf{0.18}& \textbf{0.22}& \textbf{0.44}& \textbf{0.51}& \textbf{0.48}& \textbf{0.52}& \textbf{0.18}& \textbf{0.31} &-&-&-&-&-
            \\
        \midrule
            Q2B
            & KGReasoning & \textbf{3.12$\times$} & 5.99 & 5.92 & 6.00 & 6.09 & 6.04 & 6.03 & 6.02 & 9.75 & 9.76  &-&-&-&-&-
            \\
            & KGCompiler & & \textbf{2.12} & \textbf{2.11} & \textbf{2.14} & \textbf{2.17} & \textbf{2.15} & \textbf{2.18} & \textbf{2.18} & \textbf{2.26} & \textbf{2.30} &-&-&-&-&-
            \\
        \midrule
            BetaE
            & KGReasoning & \textbf{5.28$\times$} & 34.27 & 34.93 & 34.42 & 35.20 & 33.99 & 34.61 & 34.55 & 42.15 & 42.83 & 36.93 & 36.74 & 36.91 & 36.75 & 36.88 
            \\
            & KGCompiler &  & \textbf{6.75} & \textbf{6.80} & \textbf{6.83} &\textbf{6.86} & \textbf{6.88} & \textbf{6.83} & \textbf{6.92} & \textbf{7.28} & \textbf{7.34} & \textbf{6.88} & \textbf{6.74} &\textbf{6.83} & \textbf{6.91} &\textbf{6.86}
            \\
        \midrule
            LogicE
            & KGReasoning & \textbf{1.49$\times$} & 2.4 & 2.39 & 2.79 & 3.38 & 3.83 & 3.7 & 3.96 & 2.68 & 3.36 & 3.47 & 4.14 & 4.07 & 3.83 & 3.95
            \\
            & KGCompiler &  & \textbf{1.83} & \textbf{1.86} & \textbf{1.8} & \textbf{2.25} & \textbf{2.48} & \textbf{2.49} & \textbf{2.46} & \textbf{2.07} & \textbf{2.15} & \textbf{2.2} & \textbf{2.47} & \textbf{2.45} & \textbf{2.43} & \textbf{3.35}
            \\
        \midrule
            ConE
            & KGReasoning & \textbf{8.26$\times$} & 39.9 & 40.47 & 40.77 & 41.56 & 42.13 & 41.85 & 42.06 & 70.36 & 70.81 & 41.96 & 42.56 & 42.58 & 42.36 & 42.44
            \\
            & KGCompiler &  & \textbf{3.43} & \textbf{2.48} & \textbf{2.64} & \textbf{8.69} & \textbf{8.81} & \textbf{8.82} & \textbf{9.01} & \textbf{3.95} & \textbf{4.18} & \textbf{10.76} & \textbf{11.85} & \textbf{11.75} & \textbf{11.27} & \textbf{11.9}
            \\
        \midrule
            Query2Triple
            & KGReasoning & \textbf{1.04$\times$} & 4.30 & 11.66 & 12.04 & 11.89 & 11.93 & 11.84 & 11.69 & 9.21 & 25.26 & 11.89 & 11.67 & 11.82 & 12.10 & 11.78
            \\
            & KGCompiler &  & \textbf{4.29}& \textbf{11.28}& \textbf{11.27}& \textbf{11.34}& \textbf{11.33}& \textbf{11.30}& \textbf{11.35}& \textbf{8.94}& \textbf{23.69}& \textbf{11.46}& \textbf{11.56}& \textbf{11.41}& \textbf{11.44}& \textbf{11.53}
            \\
        \bottomrule
    \end{tabular}}
    
    \caption{Comparison of computation time across different frameworks for various datasets, models, and tasks. \textbf{AVG$_{speedup}$} denotes the average speedup ratio. The computation time, from 1p to pni, is measured in milliseconds (ms).}
    \label{tab:timeres}
\end{table*}

\begin{figure*}[h]
\centering
\includegraphics[width=1\textwidth]{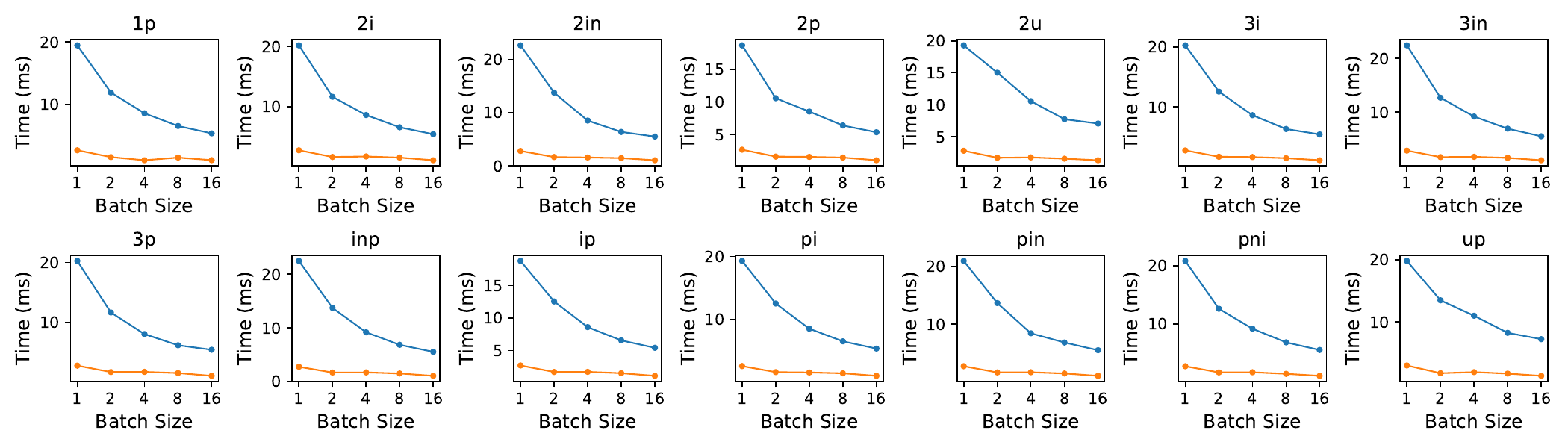}
\caption{Take the BetaE algorithm as an example, under each task type and different batch sizes, the time of processing a single query using KGReasoning (blue line) and KGCompiler (orange line) is measured respectively.}
\label{time}
\end{figure*}

\begin{figure*}[h]
    \centering
    \includegraphics[width=1\textwidth]{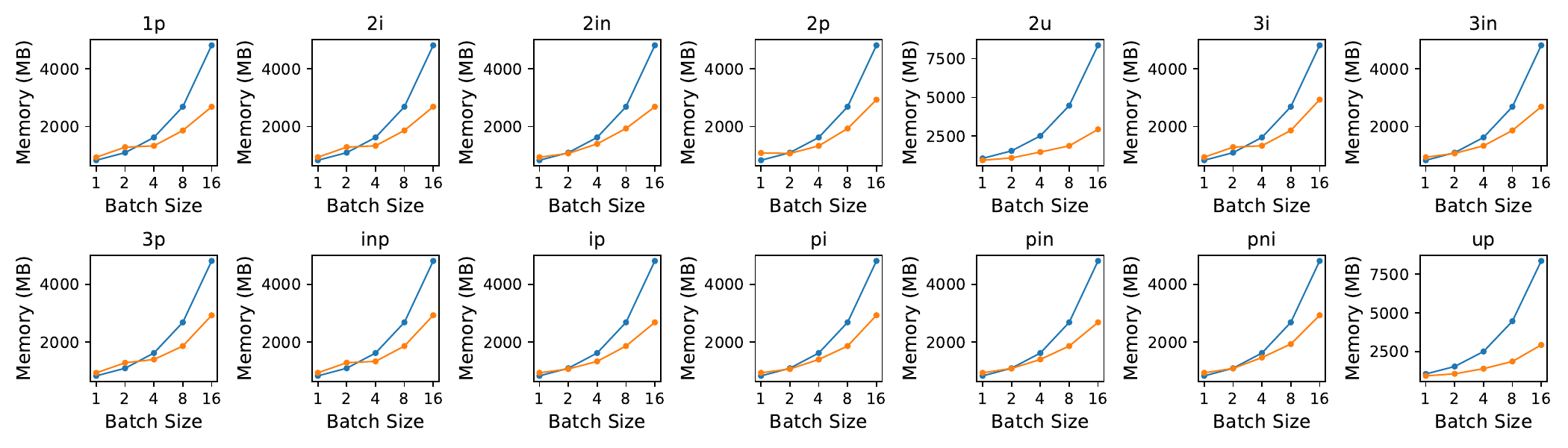}
    \caption{Take the BetaE algorithm as an example, the maximum memory footprint during computation is measured for KGReasoning (blue line) and KGCompiler (orange line) under each task type and different batch sizes.}
    \label{avgmem}
\end{figure*}

\section{Experiments}
In this section, we evaluate the performance of \textit{KGCompiler} by addressing the following three key research questions: \textbf{RQ1:} What is the speedup achieved by \textit{KGCompiler} compared to the original KG models on CLQA tasks? \textbf{RQ2:} How much memory footprint does \textit{KGCompiler} reduce? \textbf{RQ3:} Does the optimization introduced by \textit{KGCompiler} impact the accuracy of CLQA tasks?

\subsection{Experimental Setup}
\paragraph{Datasets \& Tasks.} As mentioned in \cite{NEURIPS2020_e43739bb}, FB15K \cite{bordes2013translating},  FB15K-237 \cite{toutanova2015observed}, and NELL \cite{xiong2017deeppath} are three standard datasets for CLQA tasks. To ensure a fair comparison, we select 14 typical FOL query tasks presented by \cite{NEURIPS2020_e43739bb} as our evaluation workloads, including 9 Existential Positive First Order (EPFO) queries ($1p$, $2p$, $3p$, $2i$, $3i$, $pi$, $ip$, $2u$, $up$) and 5 queries involving negation ($2in$, $3in$, $inp$, $pin$, $pni$). \par 

\paragraph{Baselines.} To evaluate the effectiveness of \textit{KGCompiler} in improving the reasoning performance of KG models, we choose 6 classic CLQA algorithms supported by the most representative multi-hop query framework KGReasoning \cite{kgreasoning}: GQE \cite{hamilton2018embedding}, Q2B \cite{ren2020query2box}, BetaE \cite{NEURIPS2020_e43739bb}, CQD \cite{arakelyan2020complex}, LogicE \cite{LogicE}, ConE \cite{ConE}, and Query2Triple \cite{Q2T}. It is worth noting that Q2B and GQE only support the $i$, $p$, and $u$ operators and do not support the $n$ operator, limiting them to completing only 9 EPFO tasks. In contrast, the other algorithms support all four operators and are capable of completing all 14 CLQA tasks. \par

\paragraph{Metrics.} We use the reasoning time of KG models as the primary performance metric, which includes the time for logical operator execution and entity similarity query. The time required for KG models to perform these tasks was measures using \textit{KGCompiler} and KGReasoning, respectively. The final result is based on the average of the times taken across 5 rounds of measurement. Furthermore, to ensure that \textit{KGCompiler} does not affect the accuracy of the KG models, we compare the consistency of Mean Reciprocal Rank (MRR) results when the KG models are executed both with or without the \textit{KGCompiler}. \par

\paragraph{Environment.} Our primary focus is on optimizing CLQA tasks executed on a single GPU, as this represents the most common use case. We provide detailed evaluation results obtained using a single NVIDIA RTX 4090 GPU, equipped with 24,564 MB of memory, and running on CUDA Toolkit 12.0, Python 3.10, and PyTorch 2.3. \par

\subsection{Main Results (RQ1)}
Table \ref{tab:timeres} shows that \textit{KGCompiler} improves the reasoning performance of all six KG models across 14 CLQA tasks at a batch size of 1, achieving up to 8.26$\times$ relative improvement over KGReasoning, with average improvements of 3.58$\times$, 3.23$\times$, and 3.96$\times$ on FB15K-237, FB15K, and NELL, respectively. Additionally, as shown in Figure \ref{time}, \textit{KGCompiler} further improves performance as the batch size increases, demonstrating its effectiveness and universality. \par

The primary factors contributing to performance acceleration are: (1) reducing the number of generated GPU kernels by merging kernels through operator fusion, which lowers the overhead associated with kernel launches and execution; (2) enhancing data reusability by loading data into faster registers, reducing slow memory access overhead; (3) improving computational parallelism by leveraging hardware resources to enable parallel computation of multiple query paths. \par

\begin{figure}[!h]
    \centering
    \includegraphics[width=80mm]{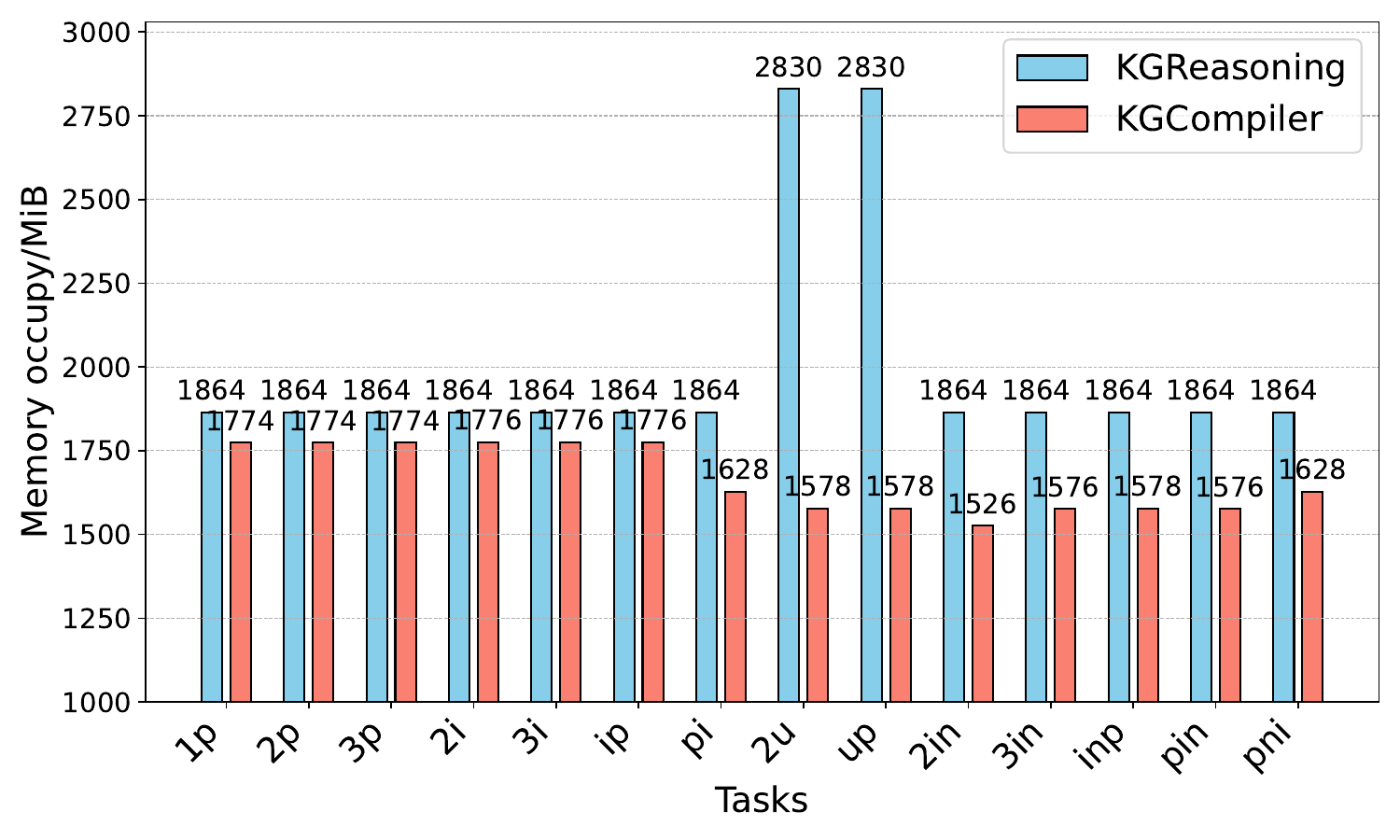}
    \caption{Take the BetaE algorithm as an example, this bar chart compares the memory usage of KGReasoning and KGCompiler, across various tasks labeled on the x-axis.}
    \label{memory}
    \vspace{-5mm}
\end{figure}

\subsection{Memory Usage (RQ2)}
There is usually a trade-off between the memory footprint and the reasoning performance for the KG models. When the batch size is large, the degree of parallelism of reasoning increases, and the total execution time decreases, but it consumes more memory space. When the batch size is small, the memory footprint is reduced, but the parallelism of the model execution is reduced, resulting in an increase in the total execution time. \par

By analyzing the data in Figure~\ref{avgmem} and Figure~\ref{memory}, when batch size is 1, we can see that \textit{KGCompiler} uses less memory in all CLQA tasks compared to KGReasoning, and the memory savings of \textit{KGCompiler} will become more significant as the batch size increases, making it possible to reasoning with a larger batch size than before with limited memory. \par

\subsection{Accuracy Measure (RQ3)}
\begin{figure}[!h]
    \centering
    \includegraphics[width=80mm]{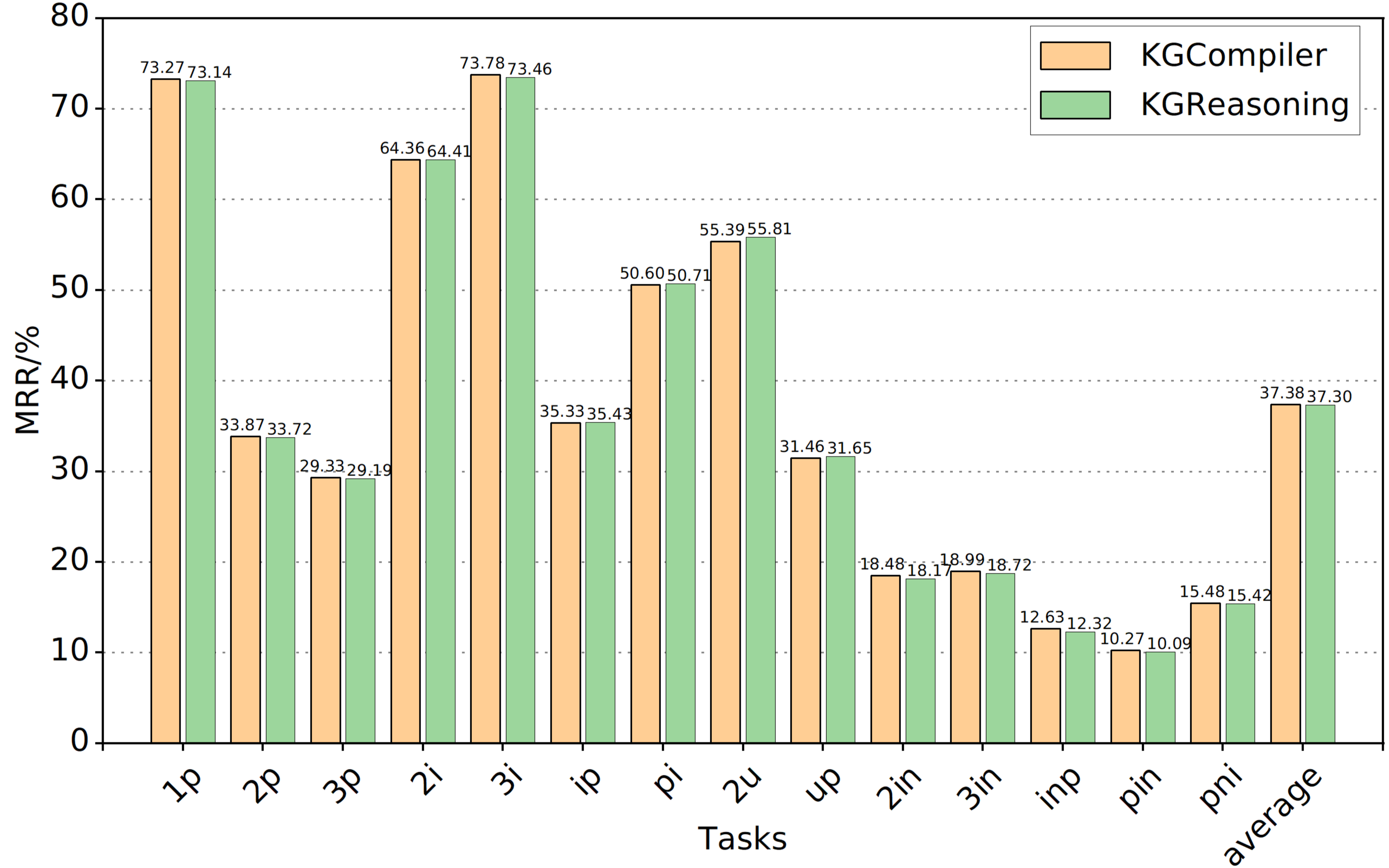}
    \caption{Take the ConE algorithm as an example, this bar chart compares the MRR of KGReasoning and KGCompiler, across various tasks labeled on the x-axis.}
    \label{mrr}
\end{figure}

As shown in Figure \ref{mrr}, the MRR values from both \textit{KGCompiler} and KGReasoning are nearly identical, indicating no significant difference. This demonstrates that \textit{KGCompiler} accelerates the reasoning process without compromising model accuracy, which is crucial for deploying efficient and accurate CLQA models in real-world applications. \par

Further research should focus on optimizing the speed and memory footprint of KG models, while maintaining their accuracy. The success of \textit{KGCompiler} demonstrates that strike a balance between computational efficiency and accuracy is achievable, providing a solid foundation for developing efficient and effective KG reasoning systems. \par

\section{Conclusion}
In this paper, we present \textit{KGCompiler}, the first KG-oriented deep learning compiler tailored for CLQA tasks. By leveraging KG-specific compilation optimizations, \textit{KGCompiler} improves the reasoning performance and reduces the memory footprint of KG models with scarcely altering their accuracy, showcasing its effectiveness and efficiency. Moreover, \textit{KGCompiler} is applicable to a variety of emerging KG models and query tasks, demonstrating its scalability and flexibility. \par

\appendix

\bibliographystyle{named}
\bibliography{ijcai25}

\end{document}